\def\paperlanguage{}
\pdfoutput=1 

\documentclass[letterpaper, 10 pt, conference]{ieeeconf}  

\usepackage{bm}
\usepackage{cite}
\usepackage{flushend}
\include{preamble}

\newcommand{\ctext}[1]{\raise0.2ex\hbox{\textcircled{\scriptsize{#1}}}}

\IEEEoverridecommandlockouts                              

\title{\LARGE \textbf
  {
    \switchlanguage%
    {%
      Daily Assistive View Control Learning of Low-Cost Low-Rigidity Robot via Large-Scale Vision-Language Model
    }%
    {%
      大規模視覚-言語モデルによる低コスト低剛性ロボットの\\生活支援視覚制御学習
    }%
  }
}

\author{Kento Kawaharazuka$^{1}$, Naoaki Kanazawa$^{1}$, Yoshiki Obinata$^{1}$, Kei Okada$^{1}$, and Masayuki Inaba$^{1}$
  \thanks{$^{1}$ The authors are with the Department of Mechano-Informatics, Graduate School of Information Science and Technology, The University of Tokyo, 7-3-1 Hongo, Bunkyo-ku, Tokyo, 113-8656, Japan.
    {\texttt\small [kawaharazuka, kanazawa, obinata, k-okada, inaba]@jsk.t.u-tokyo.ac.jp}
  }
}
\begin{document}

\maketitle
\thispagestyle{empty}
\pagestyle{empty}

\begin{abstract}
  \switchlanguage%
  {%
    In this study, we develop a simple daily assistive robot that controls its own vision according to linguistic instructions.
    The robot performs several daily tasks such as recording a user's face, hands, or screen, and remotely capturing images of desired locations.
    To construct such a robot, we combine a pre-trained large-scale vision-language model with a low-cost low-rigidity robot arm.
    The correlation between the robot's physical and visual information is learned probabilistically using a neural network, and changes in the probability distribution based on changes in time and environment are considered by parametric bias, which is a learnable network input variable.
    We demonstrate the effectiveness of this learning method by open-vocabulary view control experiments with an actual robot arm, MyCobot.
  }%
  {%
    本研究では, 言語指示によりロボットが自身の視界を制御するシンプルな日常生活支援ロボットを開発する.
    自身の顔や手, 画面を録画したり, 遠隔で所望の場所の画像を取ったりと, 日常的に必要ないくつかのタスクを言語指示に従ってこなす.
    このロボットの構成のために, 事前学習済みの大規模視覚-言語モデルと, 低剛性で安価なロボットアームを組み合わせる.
    ロボットの身体情報と視覚情報の相関関係をニューラルネットワークを用いて確率的に学習すると同時に, 時間帯や環境変化に基づく確率分布変化を, 学習可能なネットワーク入力変数であるParametric Biasにより捉える.
    低剛性なロボットアームMyCobotの言語指示に基づく実機動作実験により, その有効性を示す.
  }%
\end{abstract}

\section{Introduction}\label{sec:introduction}
\begin{figure}[t]
  \centering
  \includegraphics[width=1.0\columnwidth]{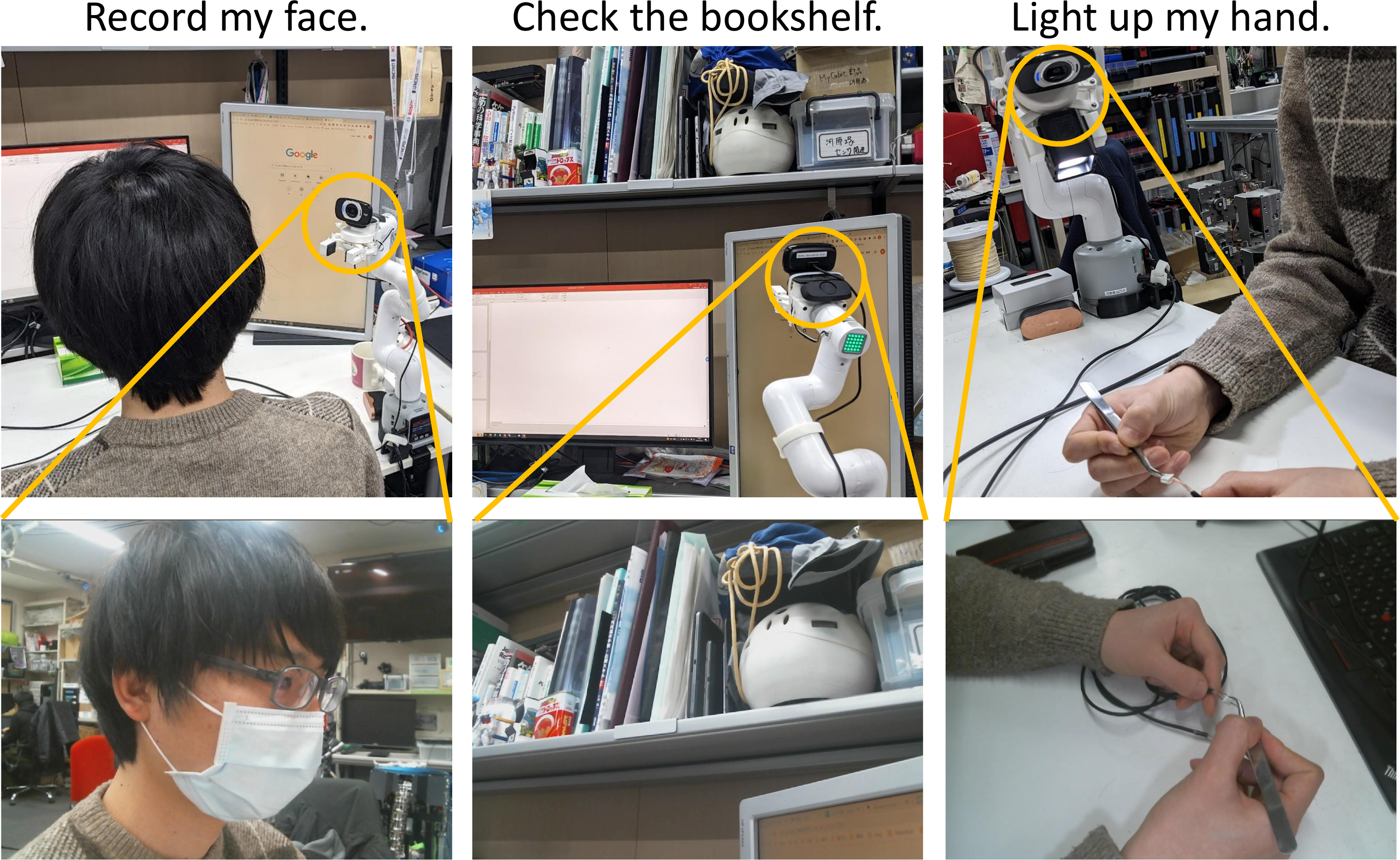}
  \vspace{-1.0ex}
  \caption{Open-vocabulary view control of a low-cost low-rigidity robot arm for daily assistive tasks. The lower figures show the image from the web camera attached to the arm-tip.}
  \vspace{-1.0ex}
  \label{figure:concept}
\end{figure}

\switchlanguage%
{%
  Robots for daily assistive tasks have been developed in various forms \cite{okada2005daily}.
  They can perform tasks such as cooking \cite{wachter2018cooking} to cleaning \cite{kim2019cleaning} and serving \cite{saito2011subwaydemo}.
  In this study, we consider a simple daily assistive robot that controls its own vision according to linguistic instructions (\figref{figure:concept}).
  The robot performs several daily assistive tasks such as recording images of the user's face, hands, and screen, remotely capturing images of desired locations, illuminating the user's hands, and so on.

  Several studies have been conducted on robotic view control.
  The most common one is on the planning task of ``Next-Best-View'' for autonomous 3D exploration of objects and environments \cite{connolly1985nbv, bircher2016nbv}.
  There are also studies on endoscope control in surgery \cite{he2021view} and gaze control in social robots \cite{zaraki2014gaze}.
  In recent years, robots exploring a 3D space to find answers to linguistic questions have been studied \cite{das2018eqa}.
  There is a study on constructing 3D maps that include language information for the navigation of mobile robots \cite{huang22vlmaps}.
  There are also some examples of pick and place tasks based on linguistic instructions \cite{hatori2018picking, shridhar2021cliport, ahn2022saycan}.
  On the other hand, this study differs from previous tasks in that it controls the robot's view in a direction appropriate to the linguistic instructions.
  While it is true that some existing methods may perform view control implicitly, this study aims to develop a system that explicitly links linguistic instructions with physical information in order to achieve more precise and intentional view control.

  In this study, we develop an open-vocabulary view control system using a low-cost low-rigidity robot arm.
  A web camera is attached to the arm-tip of MyCobot, a low-cost low-rigidity robot arm suitable for daily assistive tasks.
  The robot can perform actions based on linguistic instructions utilizing a pre-trained large-scale vision-language model that has been remarkably developed in recent years \cite{li2022largemodels, kawaharazuka2023ptvlm, kawaharazuka2023ofaga}.
  In addition, to ensure the performance of the low-cost low-rigidity robot, we introduce an experience-based learning mechanism using a neural network.
  The correlation between the visual information based on the vision-language model and the physical information of the low-cost low-rigidity robot is trained.
  In order to consider the stochastic correlation due to small changes in the visual field and the low-rigidity body, we construct a predictive model that outputs the mean and variance of sensor values.
  In addition, changes in the probability distribution of the visual information due to changes in time and environment are considered by parametric bias (PB) \cite{tani2002parametric}, which is a learnable network input variable.
  It is also possible to simply continue to collect and search images, but this would increase the cost of data management and memory, and would not capture its stochastic nature and changes in its probability distribution.
  Several experiments on actual robots show that the robot can respond to a variety of linguistic instructions and environments.

  The structure of this study is as follows.
  In \secref{sec:proposed}, we describe the construction of the probabilistic model between physical and visual information, data collection and network training, update of parametric bias representing the changes in probability distribution, and open-vocabulary view control.
  In \secref{sec:experiment}, we describe simple quantitative evaluation experiments and more practical advanced experiments.
  In \secref{sec:discussion}, we discuss the experimental results and some limitations of this study, and conclude in \secref{sec:conclusion}.
}%
{%
  日常生活支援を行うロボットはこれまで様々な形で開発されてきた\cite{okada2005daily}.
  それらは調理\cite{wachter2018cooking}から掃除\cite{kim2019cleaning}, 配膳\cite{saito2011subwaydemo}などを行うことが可能である.
  その中でも本研究では, ロボットが言語指示に従い自身の視界を制御するという非常にシンプルな日常生活支援ロボットを考える(\figref{figure:concept}).
  これは, ユーザの顔や手, 画面等を映して録画したり, 遠隔で所望の場所の画像を撮影したり, ライトで照らしたりと, 言語指示に従い日常的に必要ないくつかの補助タスクをこなす.

  これまで, 視界制御についてはいくつかの研究が行われてきた.
  最も一般的なものは, 物体や環境の自律的な三次元探索に向け, ``Next-Best-View''を計画するタスクの研究である\cite{connolly1985nbv, bircher2016nbv}.
  他にも, 手術における内視鏡の制御\cite{he2021view}やsocial robotにおけるgaze controlを行う研究もある\cite{zaraki2014gaze}.
  近年は, 言語による質問から答えを探すようにロボットが空間を探索するといった問題設定も研究されている\cite{das2018eqa}.
  移動ロボットのナビゲーションに向け, 言語が結びついた三次元マップを構築する例もある\cite{huang22vlmaps}.
  視界制御ではないが, 言語指示に基づいてpick and placeを行うような例もいくつか存在する\cite{hatori2018picking, ahn2022saycan}.
  一方で本研究はこれらのタスクとは異なり, 言語指示に適切な方向に視界を向けるものである.
  アームロボットの関節角度や関節トルクの身体情報とその手先のカメラ情報を直接結びつけ制御することを行う.
  これまでにないアプリケーションであり, 身体と言語情報の直接的な関係性を学習する点で新しい.

  本研究では, 低コスト・低剛性なロボットによるopen-vocabularyな視界制御システムを開発する.
  生活支援に適した低コスト・低剛性なアームロボットであるMyCobotの手先先端にWebカメラを取り付ける.
  近年その発展の目覚ましい事前学習済みの大規模視覚-言語モデル\cite{li2022largemodels}により言語指示に基づく動作が可能となる.
  また, 低コスト低剛性なロボットの性能を担保するため, ニューラルネットワークを用いた経験に基づく学習機構を導入する.
  視覚-言語モデルに基づく視覚情報と, 低剛性・低コストなロボットの身体情報の相関関係を学習する.
  視界の細かな変化と低剛性な身体における確率的な相関関係変化を捉えるため, 平均と分散を出力する予測モデルを構築する.
  加えて, 時間帯や環境の変化に基づくその視界情報の確率分布の変化を, 学習可能なネットワーク入力変数であるparametric bias (PB) \cite{tani2002parametric}により捉える.
  なお, 単に画像を集め続け検索をかける方法も考えられる, データ管理やメモリのコストが増加し, 確率的な性質や身体変化を捉えることはできない.
  いくつかの実機実験から, 多様な言語指示, 環境にロボットが対応可能であること示す.

  本研究の構成は以下である.
  \secref{sec:proposed}では, 身体-視覚情報の確率的なモデル構築, データ収集とネットワーク学習, 確率分布変化を表すparametric biasの更新, 視界制御について順に述べる.
  \secref{sec:experiment}では, シンプルな定量評価実験から, より実用的な応用実験について述べる.
  \secref{sec:discussion}では, 本研究における実験結果といくつかの限界について考察し, \secref{sec:conclusion}で結論を述べる.
}%

\begin{figure*}[t]
  \centering
  \includegraphics[width=2.0\columnwidth]{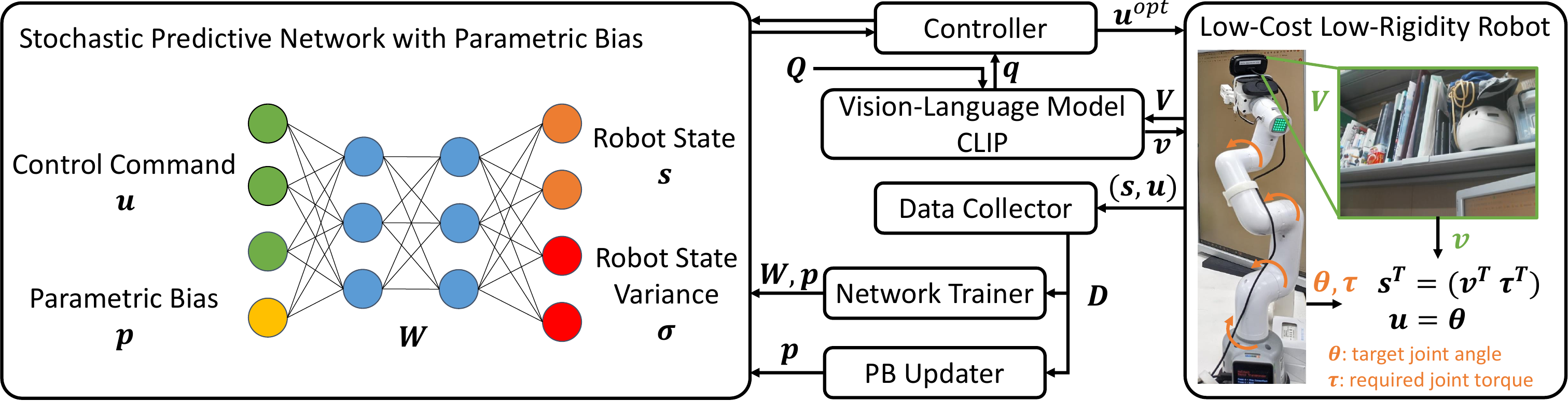}
  \vspace{-0.5ex}
  \caption{The system overview including Vision-Language Model CLIP, Data Collector, Network Trainer, PB Updater, and Controller.}
  \vspace{-1.5ex}
  \label{figure:system-overview}
\end{figure*}

\section{View Control Learning of Low-Cost Low-Rigidity Robot via Large-Scale Vision-Language Model} \label{sec:proposed}
\switchlanguage%
{%
  The overall system of this study is shown in \figref{figure:system-overview}.
  We call our network Stochastic Predictive Network with Parametric Bias (SPNPB).
  Information from the Vision-Language Model (VLM) and the robot's physical information are collected by Data Collector.
  SPNPB is trained by Network Trainer, and the Parametric Bias (PB) is updated online by Network Updater.
  SPNPB is used by Controller for open-vocabulary view control.
  Based on the network of \cite{kawaharazuka2021fetch}, time-series information is removed and a large-scale vision-language model is applied.

  The setup of MyCobot, a low-cost low-rigidity robot used in this study, is shown in the right figure of \figref{figure:system-overview}.
  A web camera is attached at the end of the arm to obtain an image $\bm{V}$.
  From the arm, the joint angle $\bm{\theta}$ can be obtained (the arm has six degrees of freedom, but only the first four are used in this study with limited angle ranges: \{[-165, 165], [-45, 45], [-22.5, 0], [-22.5, 0]\} [deg]).
  Although MyCobot cannot measure the joint torque, the necessary torque $\bm{\tau}$ can be calculated from the current joint angle $\bm{\theta}$ through its geometric model.
}%
{%
  本研究の全体システムを\figref{figure:system-overview}に示す.
  本研究のネットワークをStochastic Predictive Network with Parametric Bias (SPNPB)と呼ぶ.
  Vision-Language Model (VLM)からの情報とロボットの身体情報をData Collectorが収集する.
  Network TrainerによりSPNPBを訓練し, Network UpdaterがParametric Bias (PB)をオンラインで更新する.
  このSPNPBを用いてControllerが視界制御を行う.
  なお, 本ネットワーク構成は\cite{kawaharazuka2021fetch}から時系列情報を排除し, 大規模視覚-言語モデルを接続した形であるとも言える.

  本研究で用いる低コスト低剛性ロボットMyCobotのセットアップを\figref{figure:system-overview}に示す.
  アームの先端にはWebカメラが取り付けられており, 画像$\bm{V}$を得ることができる.
  アームからは関節角度$\bm{\theta}$を得ることができる(6自由度であるが, 本研究ではそのうち4自由度のみを角度を制限して用いる).
  MyCobotは関節トルクを測定することはできないが, 現在の関節角度$\bm{\theta}$から幾何モデルを通して必要なトルク$\bm{\tau}$を計算することができる.
}%

\subsection{Stochastic Neural Network with Parametric Bias} \label{subsec:network-structure}
\switchlanguage%
{%
  SPNPB can be expressed by the following formula,
  \begin{align}
    (\bm{s}, \bm{\sigma}) = \bm{h}(\bm{u}, \bm{p}) \label{eq:spnpb}
  \end{align}
  where $\bm{s}$ is the sensor state, $\bm{\sigma}$ is the variance of $\bm{s}$ under the assumption of Gaussian distribution, $\bm{u}$ is the control input, $\bm{p}$ is the parametric bias \cite{tani2002parametric}, and $\bm{h}$ expresses SPNPB.
  In this study, $\bm{s}$ denotes $\begin{pmatrix}\bm{v}^{T} & \bm{\tau}^{T}\end{pmatrix}^{T}$ with a vector $\bm{v}$ $(\in \mathbb{R}^{512})$ where the current image $\bm{V}$ is transformed by a vision-language model CLIP \cite{radford2021clip}, and $\bm{\tau}$ denotes the joint torque required for the gravity compensation ($\in \mathbb{R}^{4}$).
  For $\bm{u}$, the target joint angle $\bm{\theta}$ ($\in \mathbb{R}^{4}$) is used.
  Note that the values of $\bm{s}$ and $\bm{u}$ are normalized using all obtained data points.
  Since $\bm{\sigma}$ represents the variance and must always be positive, the network outputs $\bm{\sigma}$ through exponential function.
  $\bm{p}$ is responsible for representing the change in the probabilistic distribution due to changes in time and environment, and is assumed to be two-dimensional in this study.
  SPNPB consists of four fully-connected layers.
  The number of units is set to \{$N_u+N_p$, 100, 300, 500, $2N_s$\} ($N_{\{u, s, p\}}$ is the number of dimensions of $\{\bm{u}, \bm{s}, \bm{p}\}$).
  The activation function is hyperbolic tangent, and the update rule is Adam \cite{kingma2015adam}.
}%
{%
  SPNPBは数式で以下のように表せる.
  \begin{align}
    (\bm{s}, \bm{\sigma}) = \bm{h}(\bm{u}, \bm{p}) \label{eq:spnpb}
  \end{align}
  ここで, $\bm{s}$はセンサ状態, $\bm{\sigma}$はガウス分布を仮定した場合の$\bm{s}$の分散, $\bm{u}$は制御入力, $\bm{p}$はparametric bias \cite{tani2002parametric}, $\bm{h}$はSPNPBを表す.
  本研究で$\bm{s}$は, 得られた現在の画像$\bm{V}$を大規模言語モデルであるCLIP \cite{radford2021clip}によりベクトルに変換した$\bm{v}$ $(\in \mathbb{R}^{512})$と, 関節トルク$\bm{\tau}$ ($\in \mathbb{R}^{4}$)を用いる.
  $\bm{u}$は関節角度$\bm{\theta}$ ($\in \mathbb{R}^{4}$)を用いる.
  なお, $\bm{s}$, $\bm{u}$の値は得られた全データを使って正規化されている.
  また, $\bm{\sigma}$は分散を表し常に正の値を取るため, ネットワーク構造についてはexponentialを通して出力される.
  $\bm{p}$は時間帯や環境の変化によるネットワーク全体の変化を表現することを担っており, 本研究では2次元としている.

  本研究においてSPNPBは4層の全結合層からなる.
  ユニット数については, \{$N_u+N_p$, 100, 300, 500, $2N_s$\}とした(なお, $N_{\{u, s, p\}}$は$\{\bm{u}, \bm{s}, \bm{p}\}$の次元数とする).
  activation functionはhyperbolic tangent, 更新則はAdam \cite{kingma2015adam}とした.
}%

\subsection{Training of SPNPB} \label{subsec:network-training}
\switchlanguage%
{%
  We collect a dataset of $\bm{s}$ and $\bm{u}$ by random robot motions.
  By collecting data at different times of the day and in different environments, these differences can be embedded in the parametric bias as implicit information so that various data points with different distributions can be represented in a single model.
  In a series of motions in a trial $k$ in the same environment, the dataset $D_k=\{(\bm{s}^{k}_1, \bm{u}^{k}_{1}), (\bm{s}^{k}_2, \bm{u}^{k}_2), \cdots, (\bm{s}^{k}_{N_{k}}, \bm{u}^{k}_{N_{k}})\}$ is collected ($1 \leq k \leq K$, where $K$ is the total number of trials and $N_{k}$ is the number of data points for the trial $k$).
  Then, we create the dataset $D_{train}=\{(D_1, \bm{p}_1), (D_2, \bm{p}_2), \cdots, (D_{K}, \bm{p}_K)\}$ for training.
  $\bm{p}_k$ ($1 \leq k \leq K$) is the parametric bias for the trial $k$, which is a variable with a common value during the trial but a different value for other trials.
  The dataset $D_{train}$ and the following loss function are used to train SPNPB,
  \begin{align}
    P(s^{k}_{i, n}|D_{k, n}, W, \bm{p}_{k}) &= \frac{1}{\sqrt{2\pi \hat{\sigma}_{i, n}}}\exp{\left(-\frac{(\hat{s}^{k}_{i, n}-s^{k}_{i, n})^{2}}{2\hat{\sigma}_{i, n}}\right)} \label{eq:prob-density}\\
    L_{likelihood}(W, \bm{p}_{1:K}|D_{train}) &= \prod^{K}_{k=1}\prod^{N_{k}}_{n=1}\prod^{N_{s}}_{i=1}P(s^{k}_{i, n}|D_{k, n}, W, \bm{p}_{k}) \label{eq:likelihood}\\
    L_{train} &= -\log(L_{likelihood}) \label{eq:train-loss}
  \end{align}
  where $P$ is the probability density function, $\{s, \sigma\}_{i}$ is $\{s, \sigma\}$ of the $i$-th sensor, $D_{k, n}$ is the $n$-th data point in $D_k$, $W$ is the network weight of SPNPB, $\{\hat{s}, \hat{\sigma}\}$ is the value of $\{s, \sigma\}$ predicted from SPNPB using the dataset $D_{k, n}$, the current weight $W$, and the current parametric bias $\bm{p}_{k}$ for the trial $k$, and $\bm{p}_{1:K}$ is a vector of $\bm{p}_{k}$ within $1 \leq k \leq K$.
  $L_{likelihood}$ denotes the likelihood function for $W$ and $\bm{p}$ given $D_{train}$, and we consider the problem of maximizing it.
  This function is a modification of the loss function in \cite{murata2013stochastic}.
  We simplify the computation to the summation of $\log(P)$ by performing the transformation as in \equref{eq:train-loss}, making it the problem of minimizing $L_{train}$.
  In the usual training, only the network weight $W$ is updated, but in this study, $W$ and $\bm{p}_{k}$ are updated at the same time.
  Note that each $\bm{p}_{k}$ is optimized with an initial value of $\bm{0}$.
}%
{%
  ランダムなロボットの動作によって$\bm{s}$と$\bm{u}$のデータを収集する.
  この際, 動作させる時間帯や環境を変化させながらデータを収集することで, それらを暗黙的な情報としてparametric biasに埋め込むことができる.
  ある同一の環境における一連の試行$k$においてデータ$D_k=\{(\bm{s}^{k}_1, \bm{u}^{k}_{1}), (\bm{s}^{k}_2, \bm{u}^{k}_2), \cdots, (\bm{s}^{k}_{N_{k}}, \bm{u}^{k}_{N_{k}})\}$を得る($1 \leq k \leq K$, $K$は全試行回数, $N_{k}$はその試行$k$に関するデータ数とする).
  そして, 学習に用いるデータ$D_{train}=\{(D_1, \bm{p}_1), (D_2, \bm{p}_2), \cdots, (D_{K}, \bm{p}_K)\}$を得る.
  $\bm{p}_k$はその試行$k$に関するparametric biasであり, その試行中については共通の値で, 異なる試行については別の値となる変数である.
  このデータ$D_{train}$と以下の損失関数を用いてSPNPBを学習させる.
  \begin{align}
    P(s^{k}_{i, n}|D_{k, n}, W, \bm{p}_{k}) &= \frac{1}{\sqrt{2\pi \hat{\sigma}_{i, n}}}\exp{\left(-\frac{(\hat{s}^{k}_{i, n}-s^{k}_{i, n})^{2}}{2\hat{\sigma}_{i, n}}\right)} \label{eq:prob-density}\\
    L_{likelihood}(W, \bm{p}_{1:K}|D_{train}) &= \prod^{K}_{k=1}\prod^{N_{k}}_{n=1}\prod^{N_{s}}_{i=1}P(s^{k}_{i, n}|D_{k, n}, W, \bm{p}_{k}) \label{eq:likelihood}\\
    L_{train} &= -\log(L_{likelihood}) \label{eq:train-loss}
  \end{align}
  ここで, $P$は確率密度関数, $\{s, \sigma\}_{i}$は$i$番目のセンサの$\{s, \sigma\}$, $D_{k, n}$は$D_k$における$n$番目のデータ, $W$はSPNPBのネットワークの重み, $\{\hat{s}, \hat{\sigma}\}$はデータ$D_{k, n}$, 現在の重み$W$, 試行$k$に関する現在のparametric bias $\bm{p}_{k}$を使ってSPNPBから予測された$\{s, \sigma\}$の値, $\bm{p}_{1:K}$は$1 \leq k \leq K$の$\bm{p}_{k}$をまとめたベクトルを表す.
  $L_{likelihood}$は $D_{train}$が与えられたときの$W$と$\bm{p}$に関する尤度関数を表し, これを最大化する問題を考える.
  この損失関数は\cite{murata2013stochastic}を本研究の形のネットワークに応用した形と言える.
  このとき, \equref{eq:train-loss}のように変換を行うことで$\log(P)$の足し算へと計算を簡単にし, $L_{train}$の最小化問題に変形する.
  通常の学習ではネットワークの重み$W$のみを更新するが, 本研究では$W$と$\bm{p}_{k}$を同時に更新していく.
  なお, 全$\bm{p}_k$は初期値を0として最適化されている.
}%

\subsection{Update of Parametric Bias} \label{subsec:network-update}
\switchlanguage%
{%
  By continuously updating the parametric bias $\bm{p}$, the robot can always adapt to changes in time and environment.
  While the robot is moving, we always collect the dataset $D_{update}$ of $\bm{s}$ and $\bm{u}$.
  The update of $\bm{p}$ starts when the number of the collected data $N^{update}_{data}$ exceeds the threshold $N^{update}_{thre}$.
  Data exceeding $N^{update}_{max}$ is discarded from the oldest ones.
  We use \equref{eq:prob-density} -- \equref{eq:train-loss} as a loss function, and train SPNPB with $N^{update}_{batch}$ and $N^{update}_{epoch}$ as the number of batches and epochs, respectively.
  Here, the network weight $W$ is fixed and only $\bm{p}$ is updated.
  By updating only $\bm{p}$, we can avoid catastrophic forgetting and overfitting while adapting to the current environment.
  In this study, we set $N^{update}_{batch}=N^{update}_{data}$, $N^{update}_{thre}=100$, $N^{update}_{max}=200$, and $N^{update}_{epoch}=3$, and use Momentum SGD \cite{qian1999momentum} as the update rule.
}%
{%
  parametric bias $\bm{p}$を更新し続けることで, 時間帯や環境の変化に常に適応することができる.
  ロボットが動いている際に, 常に$\bm{s}$と$\bm{u}$のデータ$D_{update}$を取得しておく.
  この$D_{update}$のデータ数$N^{update}_{data}$が閾値$N^{update}_{thre}$を超えたところから$\bm{p}$の更新を開始する.
  $N^{update}_{max}$を超えたデータは古いものから破棄する.
  損失関数は\equref{eq:prob-density} -- \equref{eq:train-loss}を使用し, バッチ数を$N^{update}_{batch}$, $N^{update}_{epoch}$として学習を行う.
  この際, ネットワークの重み$W$は固定し, $\bm{p}$のみを更新する.
  $\bm{p}$のみを更新することで, 破壊的忘却や過学習を避けながら適応することができる.
  なお, 本研究では$N^{update}_{batch}=N^{update}_{data}$, $N^{update}_{thre}=100$, $N^{update}_{max}=200$, $N^{update}_{epoch}=3$, 更新則はMomentum SGD \cite{qian1999momentum}として学習を行う.
}%

\begin{figure}[t]
  \centering
  \includegraphics[width=0.95\columnwidth]{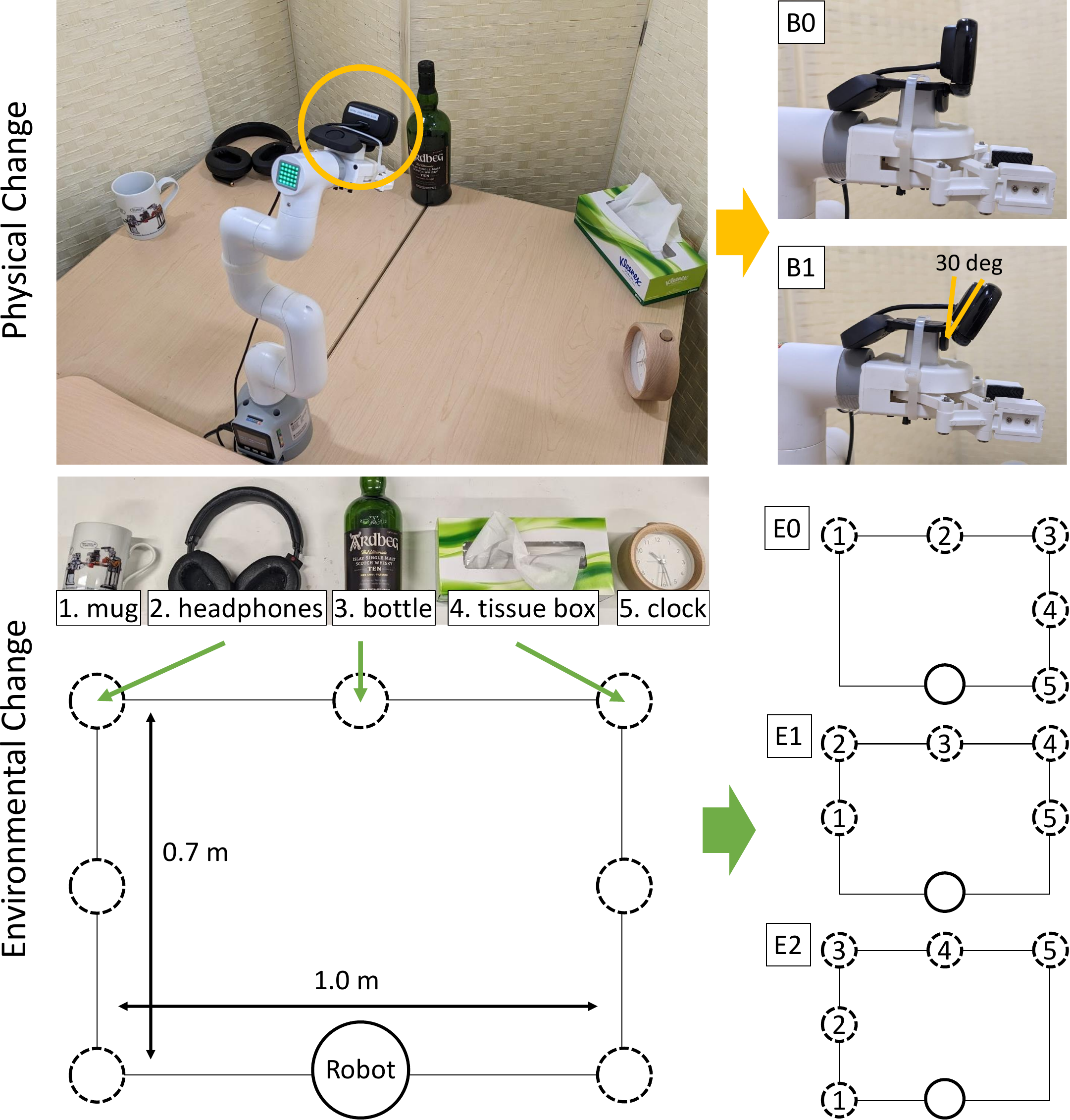}
  \vspace{-0.5ex}
  \caption{The setup of the basic experiment. The upper figures show the changes in physical state (two changes in the angle of the web camera attached to the tip of the robot arm) and the lower figure shows the changes in environmental state (three changes in the arrangement of the five target objects).}
  \vspace{-1.5ex}
  \label{figure:basic-setup}
\end{figure}

\begin{figure}[t]
  \centering
  \includegraphics[width=0.8\columnwidth]{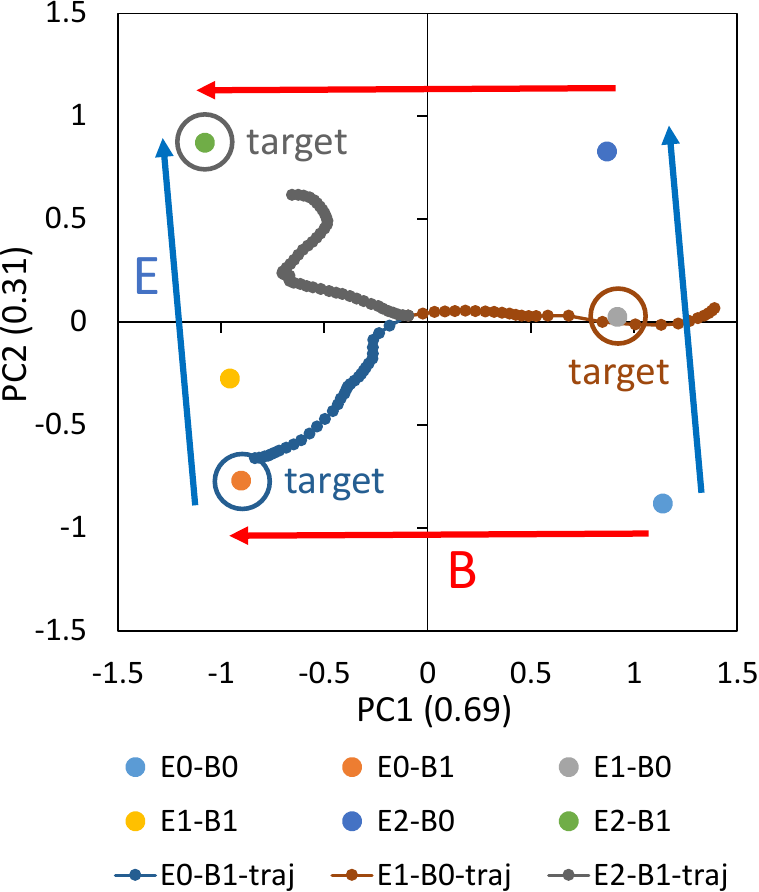}
  \vspace{-0.5ex}
  \caption{The arrangement of the trained parametric bias and its trajectory during the online update of parametric bias regarding three environmental and physical states in the basic experiment.}
  \vspace{-1.5ex}
  \label{figure:basic-pb}
\end{figure}

\subsection{Open-Vocabulary View Control} \label{subsec:view-control}
\switchlanguage%
{%
  By using the trained SPNPB, it is possible to control the robot view from linguistic instructions.
  First, we prepare a linguistic instruction $Q$, and this is transformed into a latent vector $\bm{q}$ by using CLIP which is in the same latent space as $\bm{v}$.
  Next, the initial value $\bm{u}^{init}$ of the control input $\bm{u}^{opt}$ to be optimized is determined, and the control input $\bm{u}^{opt}$ is updated by repeating the following process,
  \begin{align}
    L_{control} &= -\bm{q}\cdot\hat{\bm{v}} + C_{\tau}||\hat{\bm{\tau}}||_{2}\label{eq:control-loss}\\
    \bm{u}^{opt} &\gets \bm{u}^{opt} - \gamma\frac{\partial L_{control}}{\partial \bm{u}^{opt}}\label{eq:control-opt}
  \end{align}
  where $\{\hat{\bm{v}}, \hat{\bm{\tau}}\}$ is the mean of $\{\bm{v}, \bm{\tau}\}$ predicted from the current SPNPB and $\bm{u}^{opt}$, $C_{\tau}$ is the weight of the loss function, and $\gamma$ is the learning rate.
  The first term on the right-hand side of \equref{eq:control-loss} is the loss to obtain the control input closest to the given linguistic instruction, and the second term on the right-hand side is the loss to reduce the required torque of the posture.
  \equref{eq:control-opt} updates $\bm{u}^{opt}$ using backpropagation technique and the gradient descent method.
  Here, the initial value $\bm{u}^{init}$ uses $N^{control}_{init}$ number of random $\bm{u}$.
  By starting the optimization from various $\bm{u}^{init}$, we can avoid the solution from falling into the local minima.
  Although $\gamma$ can be a fixed value, in this study, each $\bm{u}^{opt}$ is updated by using $N^{control}_{batch}$ number of $\gamma$ which are the logarithmically divided values in $[0, \gamma^{max}]$, and then $\bm{u}^{opt}$ with the smallest loss when running \equref{eq:control-loss} is used repeatedly to obtain faster convergence.
  \equref{eq:control-loss} -- \equref{eq:control-opt} are performed $N^{control}_{epoch}$ times, and the obtained $\bm{u}^{opt}$ is sent to the actual robot.

  In this study, we set $N^{control}_{init}=30000$, $N^{control}_{batch}=100$, $N^{control}_{epoch}=2$, $\gamma^{max}=0.1$, and $C_{\tau}=0.0001$.
}%
{%
  学習されたSPNPBを用いて, 言語指示からロボットの制御を行うことが可能である.
  まず, 言語指示$Q$をCLIPを用いて$\bm{v}$と同じ潜在空間ベクトル$\bm{q}$に変換する.
  次に, 最適化する制御入力$\bm{u}^{opt}$の初期値$\bm{u}^{init}$を決定し, 以下の処理を繰り返すことで制御入力$\bm{u}^{opt}$を更新していく.
  \begin{align}
    L_{control} &= -\bm{q}\cdot\hat{\bm{v}} + C_{\tau}||\hat{\bm{\tau}}||_{2}\label{eq:control-loss}\\
    \bm{u}^{opt} &\gets \bm{u}^{opt} - \gamma\frac{\partial L_{control}}{\partial \bm{u}^{opt}}\label{eq:control-opt}
  \end{align}
  ここで, $\{\hat{\bm{v}}, \hat{\bm{\tau}}\}$は現在のSPNPBと$\bm{u}^{opt}$から予測された$\bm{v}$の平均値, $C_{\tau}$は損失関数の重み, $\gamma$は学習率である.
  \equref{eq:control-loss}の右辺第一項は与えられた言語指令に最も近い制御入力を得るための損失, 右辺第二項は動作の必要トルクを小さくするための損失である.
  \equref{eq:control-opt}は誤差逆伝播と最急降下法を用いて$\bm{u}^{opt}$を更新する.
  このとき, $\bm{u}^{init}$は, $N^{control}_{init}$個のランダムな$\bm{u}$を用いる.
  多様な$\bm{u}^{init}$から最適化を始めることで局所解に陥ることを防ぐことができる.
  また, $\gamma$は固定値でも良いが, 本研究では$[0:\gamma^{max}]$を対数的に等間隔に分けた$N^{control}_{batch}$個の$\gamma$を用いて$\bm{u}^{opt}$をそれぞれ更新し, \equref{eq:control-loss}を実行した際に最も損失の小さかった$\bm{u}^{opt}$を利用することを繰り返すことでより速い収束を得る.
  \equref{eq:control-loss} -- \equref{eq:control-opt}を$N^{control}_{epoch}$回行い, 最終的に得られた最も損失の小さな$\bm{u}^{opt}$を実機へ送る.

  なお, 本研究では$N^{control}_{init}=30000$, $N^{control}_{batch}=100$, $N^{control}_{epoch}=2$, $\gamma^{max}=0.1$, $C_{\tau}=0.0001$とした.
}%

\section{Experiments} \label{sec:experiment}
\switchlanguage%
{%
  First, we conduct a basic quantitative experiment in a small area surrounded by objects and walls.
  Next, we conduct an advanced experiment in a wider and more cluttered environment to demonstrate the effectiveness of the method.
}%
{%
  まず, 周囲を物体に囲まれた箱庭の環境で定量的な基礎実験を行う.
  次に, より広く乱雑な環境で実験を行い, その有効性を示す.
}%

\begin{figure*}[t]
  \centering
  \includegraphics[width=1.8\columnwidth]{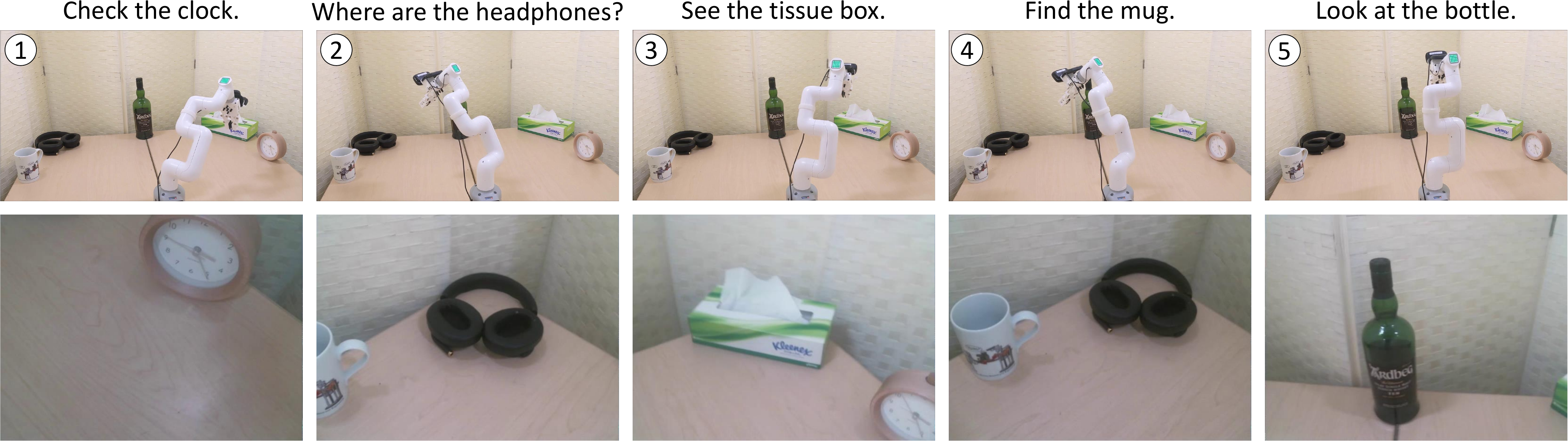}
  \vspace{-0.5ex}
  \caption{The open-vocabulary view control of the basic experiment. The lower figures show the images from the web camera attached to the arm-tip.}
  \vspace{-1.0ex}
  \label{figure:basic-exp}
\end{figure*}

\begin{figure*}[t]
  \centering
  \includegraphics[width=2.0\columnwidth]{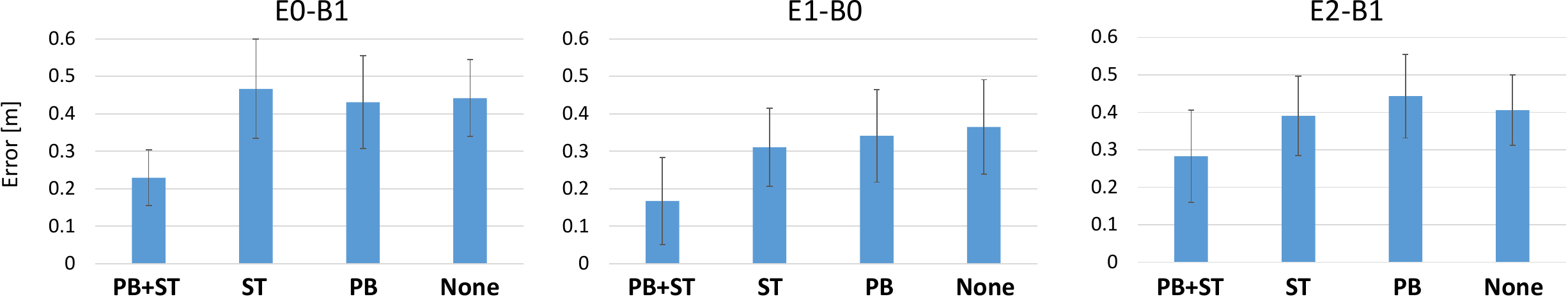}
  \vspace{-0.5ex}
  \caption{The comparison of view control errors among various neural network models: \textbf{PB+ST} - the proposed SPNPB, \textbf{ST} - stochastic predictive model without parametric bias, \textbf{PB} -the normal predictive model with parametric bias, and \textbf{None} - the general neural network without parametric bias, under three different environmental and physical states.}
  \vspace{-1.5ex}
  \label{figure:basic-graph}
\end{figure*}

\subsection{Basic Experiment} \label{subsec:basic-exp}
\switchlanguage%
{%
  The setup of the basic experiment is shown in \figref{figure:basic-setup}.
  Five objects -- 1. mug, 2. headphones, 3. bottle, 4. tissue box, and 5. clock -- are arranged in front of the robot on a desk.
  By surrounding the desk with walls, the robot can uniquely determine the direction in which it should look at to face the object corresponding to each linguistic instruction.
  Parametric bias can embed not only environmental changes but also the physical changes of a low-rigidity robot.
  In this study, the angle of the web camera attached to the tip of the robot arm is changed to 0 degrees (state B0) and 30 degrees (state B1).
  Also, we prepare three environments, E0, E1, and E2, in which the positions of the five objects are shifted one by one as shown in the lower right figure of \figref{figure:basic-setup}.

  A dataset is collected for 100 seconds for each of the six states, which are combinations of two physical changes and three environmental changes.
  The joint angles are fed randomly, and data collection is performed at 10 Hz, obtaining a total of 6000 data points.
  SPNPB is trained based on this dataset.
  The arrangement of the trained parametric bias when applying Principle Component Analysis (PCA) is shown in \figref{figure:basic-pb}.
  Note that the parametric bias for each state is expressed as E\{0, 1, 2\}-B\{0, 1\}.
  It can be seen that each PB is regularly arranged along the environmental and physical changes.
  Since no information on environmental and physical changes is given during the training, various state changes can be implicitly self-organized in the space of PB.

  Next, we conducted an experiment in which parametric bias is updated online for three states, E0-B1, E1-B0, and E2-B1.
  The trajectories ``-traj'' when updating the parametric bias from random motions are shown in \figref{figure:basic-pb}.
  The initial value of $\bm{p}$ is $\bm{0}$, and it can be seen that the parametric bias gradually approaches the appropriate value trained for the current physical and environmental conditions.
  In other words, the robot can gradually recognize the current state correctly even if its body and surrounding environment change.

  Finally, we conducted an experiment of view control using the trained SPNPB.
  The experiment is performed for the state E1-B0 after the correct PB is recognized.
  Here, linguistic instructions of ``Check the clock.'', ``Where are the headphones?'', ``See the tissue box.'', ``Find the mug.'', and ``Look at the bottle.'' are given in that order.
  The result of open-vocabulary view control is shown in \figref{figure:basic-exp}.
  It can be seen that the robot's view is correctly controlled so that the object mentioned in the linguistic instruction fits into the camera image.
  The results of comparative experiments of this view control for various physical and environmental conditions of E0-B1, E1-B0, and E2-B1, while changing the neural network model used, are shown in \figref{figure:basic-graph}.
  The models used are SPNPB of this study (\textbf{PB+ST}), SPNPB without parametric bias (\textbf{ST}), SPNPB with the loss function \equref{eq:train-loss} changed to a general mean squared error (\textbf{PB}), and a general neural network model excluding parametric bias and setting the loss function as mean squared error (\textbf{None}).
  For five target objects $O$, five linguistic instructions, ``Look at the $O$.'', ``See the $O$.'', ``Find the $O$.'', ``Check the $O$.'', and ``Where is the $O$?'' are used.
  The mean and variance of the distance between the camera's line-of-sight vector and the predefined location of the target object (the distance from a point to a line) in the 25 experiments are shown in \figref{figure:basic-graph}.
  From the results, it is found that the error of view control in \textbf{PB+ST} is the lowest, while the accuracy in \textbf{ST}, \textbf{PB}, and \textbf{None} is much lower.
  In particular, the errors for E0-B1 and E1-B0 when using \textbf{ST}, \textbf{PB}, and \textbf{None}, are more than twice as large as those when using \textbf{PB+ST}.
}%
{%
  基礎実験のセットアップを\figref{figure:basic-setup}に示す.
  ロボットの正面に1. mug, 2. headphones, 3. bottle, 4. tissue box, 5. clockの5つの物体を机の上に配置している.
  周囲を壁で囲むことで, それぞれの言語指示に対応する見るべき方向が一意に定まるセットアップとしている.
  Parametric Biasには環境変化だけでなく, 低剛性なロボットの身体変化等も埋め込むことができる.
  本研究では, ロボットアーム先端に取り付けられたウェブカメラの角度を0度とした状態B0と30度とした状態B1を用意した.
  また, 環境変化としてはこれら1-5の物体の配置が\figref{figure:basic-setup}の右下のように一箇所ずつズレたE0, E1, E2の3つの環境を用意した.

  身体変化と環境変化をかけ合わせた計6つの状態についてそれぞれ100秒間のデータ収集を行った.
  関節角度をランダムに送り, データ取得は10Hzで行い, 計6000のデータを得ている.
  このデータを元にSPNPBを学習させた際のParametric Biasに対してPrinciple Component Analysis (PCA)をかけたときの配置を\figref{figure:basic-pb}に示す.
  なお, ここではそれぞれの状態におけるParametric BiasをE\{0, 1, 2\}-B\{0, 1\}の形で表現している.
  それぞれのPBが環境変化や身体変化に沿って規則的に配置されていることがわかる.
  学習の際に環境変化や身体変化に関する情報は一切与えていないため, 多様な状態変化をPBの空間内に自己組織化可能である.

  次に, Parametric Biasをオンライン更新する実験を行った.
  E0-B1, E1-B0, E2-B1という3つの状態を作り, ランダムな動作からParametric Biasを更新した際の軌跡``-traj''を\figref{figure:basic-pb}に示す.
  $\bm{p}$の初期値は$\bm{0}$であるが, 次第にParametric Biasが現在の身体・環境状態において訓練された適切な値に近づいていっていることがわかる.
  つまり, ロボットは身体や周囲の環境が変化しても, 自身の置かれた現在状態を徐々に正しく認識することが可能である.

  最後に, SPNPBを使った視界制御の実験を行った.
  E1-B0の状態について, 正しいPBが認識できた状態において実験を行っている.
  ``Check the clock.'', ``Where are the headphones ?'', ``See the tissue box.'', ``Find the mug.'', ``Look at the bottle''の順で言語指示を変化させたときの様子を\figref{figure:basic-exp}に示す.
  言語指示に含まれる物体がカメラ画像中に収まるように正しく視界が制御出来ていることがわかる.
  この視界制御をE0-B1, E1-B0, E2-B1の多様な身体・環境状態において, 使用するモデルを変化させながら比較実験を行った際の結果を\figref{figure:basic-graph}に示す.
  モデルとしては, 本研究のSPNPB (\textbf{PB+ST}), SPNPBからPBを除いた場合(\textbf{ST}), SPNPBの損失関数\equref{eq:train-loss}を一般的なmean squared errorとした場合(\textbf{PB}), PBを除きかつ損失関数をmean squared errorとした一般的なニューラルネットワークの場合(\textbf{None})の4種類を比較している.
  5種類の指令物体$O$に対して, ``Look at the $O$'', ``See the $O$'', ``Find the $O$'', ``Check the $O$'', ``Where is the $O$ ?''の5種類の言語指示を行った.
  つまり, 25回の実験結果において, カメラの視線ベクトルと指令物体の位置の距離の平均と分散を示している.
  \figref{figure:basic-graph}から, \textbf{PB+ST}における視界制御の誤差が最も低く, それに対して\textbf{ST}や\textbf{PB}, \textbf{None}の精度は大きく下がることがわかった.
  特に, E0-B1やE1-B0では誤差に2倍以上の差が出ている.
}%

\begin{figure}[t]
  \centering
  \includegraphics[width=0.75\columnwidth]{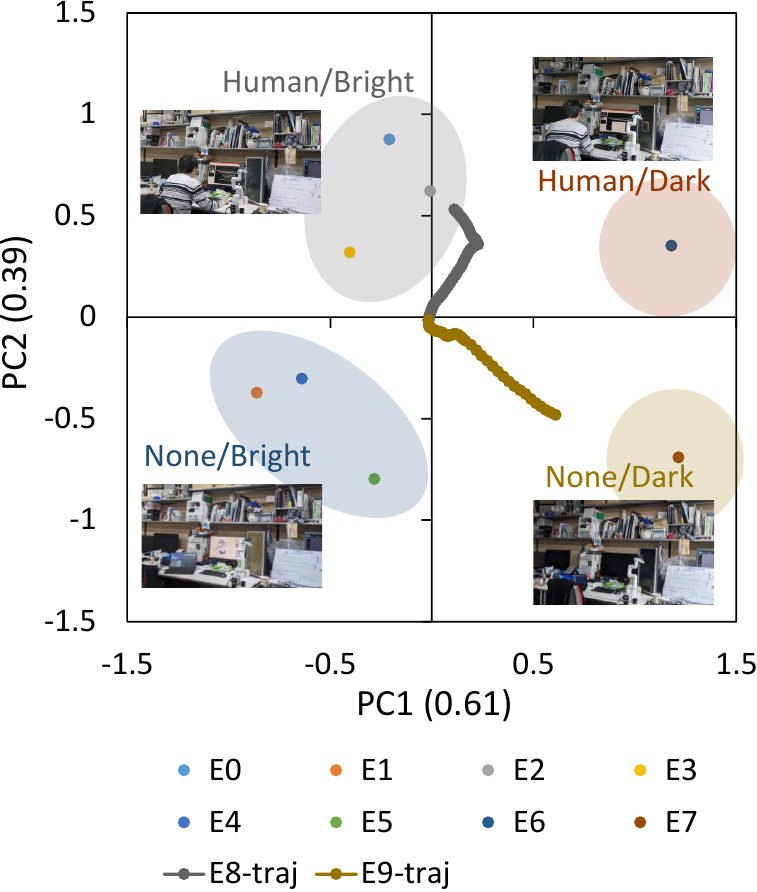}
  \vspace{-0.5ex}
  \caption{The arrangement of the trained parametric bias and its trajectory during the online update of parametric bias in the advanced experiment.}
  \vspace{-1.5ex}
  \label{figure:advanced-pb}
\end{figure}

\begin{figure*}[t]
  \centering
  \includegraphics[width=2.0\columnwidth]{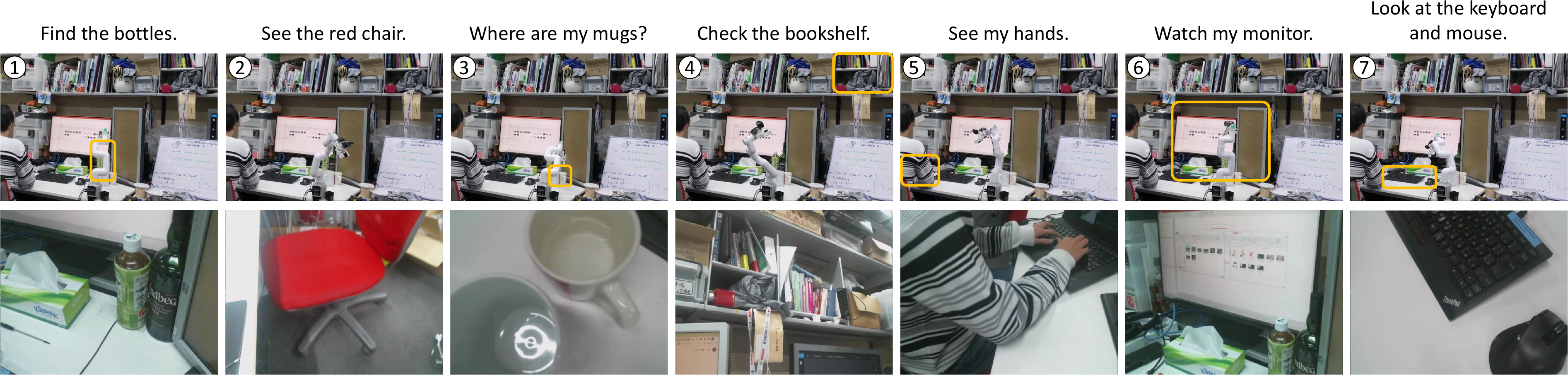}
  \vspace{-0.5ex}
  \caption{The open-vocabulary view control of the advanced experiment. The lower figures show the images from the camera.}
  \vspace{-1.5ex}
  \label{figure:advanced-exp}
\end{figure*}


\subsection{Advanced Experiment} \label{subsec:advanced-exp}
\switchlanguage%
{%
  We conducted an advanced experiment in a setting closer resembling our living space.
  A monitor, a keyboard, a mug, a bottle, a tissue case, and various other objects were placed on the desk.
  We obtained data from random motions in various environments at different times of the day.
  The environments are divided into combinations of two states: one with or without a person sitting at a desk (Human or None), and one with all lights on (Bright) or some lights on (Dark).
  Data collection is performed at 10 Hz for 100 seconds for each of eight different time periods E0--E7, obtaining a total of 8000 data points.
  E0, E2, and E3 are Human/Bright cases, E1, E4, and E5 are None/Bright cases, E6 is a Human/Dark case, and E7 is a None/Dark case.
  SPNPB is trained based on this dataset.
  The arrangement of the trained parametric bias when applying PCA, and the corresponding environmental states are shown in \figref{figure:advanced-pb}.
  It is found that each PB is regularly arranged along the environmental changes, and the space of PB is implicitly self-organized.

  Next, we conducted an experiment to update the parametric bias online.
  The trajectories ``-traj'' when updating parametric bias from random motion for a new Human/Bright environment E8 and a new None/Dark environment E9 are shown in \figref{figure:advanced-pb}.
  For both cases, we can see that $\bm{p}_{k}$ is gradually updated toward that of the same previously trained environment: in case of E8, toward $\bm{p}_{k}$ of Human/Bright environment E0, E2, and E3, and in case of E9, toward $\bm{p}_{k}$ of None/Dark environment E7.
  The same performance as that of \secref{subsec:basic-exp} can be achieved even in a cluttered environment.

  Finally, we conducted a view control experiment for the environment of E8 using the trained SPNPB and updated PB.
  Here, linguistic instructions of ``Find the bottles.'', ``See the red chair.'', ``Where are my mugs?'', ``Check the bookshelf.'', ``See my hands.'', ``Watch my monitor.'', ``Look at the keyboard and mouse.'' are given in that order.
  The result of open-vocabulary view control is shown in \figref{figure:advanced-exp}.
  Similar to \secref{subsec:basic-exp}, it is possible for the robot to view the objects and environments as indicated.
}%
{%
  より生活空間に近い場所における応用実験を行った.
  机の上にはモニターやキーボード, マグカップやボトル, ティッシュケース, その他多様な物体が置かれている.
  ここで時間帯を変えて様々な環境におけるデータをランダム動作から取得した.
  その環境は主に, 机に人が座っている時間(Human)といない時間(None), 全ての電気がついている明るい時間(Bright)と一部の電気がついていない薄暗い時間(Dark)に分けられる.
  8つの時間帯の異なる環境E0-E7それぞれについて10Hzで100秒間, 合計8000のデータを取得した.
  特に,E0, E2, E3はHuman/Bright, E1, E4, E5はNone/Bright, E6はHuman/Dark, E7はNone/Darkな環境のデータである.
  得られたデータを元にSPNPBを学習させた際のParametric Biasに対してPCAをかけたときの配置, 対応する環境状態を\figref{figure:advanced-pb}に示す.
  それぞれのPBが環境変化に沿って規則的に配置され, PBの空間が自己組織化していることがわかった.

  次に, Parametric Biasをオンライン更新する実験を行った.
  新しいHuman/Brightな環境E8とNone/Darkな環境E9について, ランダム動作からParametric Biasを更新した際の軌跡``-traj''を\figref{figure:advanced-pb}に示す.
  どちらのケースについても, E8の場合は同じHuman/BrightなE0, E2, E3の$\bm{p}_{k}$の方へ, E9の場合は同じNone/DarkなE7の$\bm{p}_{k}$の方へ$\bm{p}$が徐々に正しく更新されていくことがわかる.
  本実験のような乱雑な環境でも\secref{subsec:basic-exp}と同様の性能が発揮可能である.

  最後に, E8の環境について, 訓練されたSPNPBと更新されたPBを用いて視界制御実験を行った.
  ``Find the bottles.'', ``See the red chair.'', ``Where are my mugs?'', ``Check the bookshelf.'', ``See my hands.'', ``Watch my monitor'', ``Look at the keyboard and mouse.''の順で言語指示を変化させたときの様子を\figref{figure:advanced-exp}に示す.
  \secref{subsec:basic-exp}と同様に, 指示した通りの物体のや環境の方向をロボットが視認することが可能であった.
}%

\section{Discussion} \label{sec:discussion}
\switchlanguage%
{%
  The obtained experimental results are discussed.
  First, we conducted basic quantitative experiments in a controlled environment where the direction of the target object is uniquely determined.
  The changes in the probability distribution of the network based on changes in the body and the environment can be regularly self-organized in the space of parametric bias.
  By updating PB online, the current state of the body and the environment can be appropriately identified.
  The obtained SPNPB can then be used to direct the robot's gaze in the direction of appropriate objects based on linguistic instructions.
  It is found that this performance is achieved only when both the probabilistic predictive model and parametric bias are used, and when either one of them is not used, the control performance is reduced to about half.
  Next, we conducted advanced view control experiments in a cluttered environment that is more similar to that of daily life.
  By collecting and learning data at various times of the day with different parameters such as the presence of a human and the brightness of the room, these environmental changes become self-organized in the space of parametric bias.
  As in the basic experiment, the current environment is able to be understood appropriately, and the robot can perform view control for various objects and environments based on linguistic instructions.
  These results indicate that even with a low-cost low-rigidity body structure, and even with a large-scale vision-language model whose output changes with slight changes in the image, appropriate and adaptive view control is possible by considering the relationship between vision and body in a stochastic form, and by incorporating large changes in the form of parametric bias.

  Limitations and future prospects of this study are described.
  First, this study mainly deals with view control, and it does not actually perform a task such as recording the image in response to the command ``please record''.
  Of course, such a command can be easily implemented by recognizing the word ``record'', but there is no limit to the variety of commands, such as ``send an image via chat'', ``read the sentence out loud'', or ``turn on the lights''.
  In the future, we would like to develop this system into a more practical system that automatically uses multiple APIs and view control according to linguistic instructions, using large-scale language models.
  Second, the system currently does not accept commands such as ``a little more to the right'' or ``look at the back''.
  This is because there is no embodiment in the large-scale vision-language model itself, which is an interesting issue to be addressed in the future.
  In addition, we would like to construct a system that can always keep moving by regularly collecting data and learning networks.
  For a more practical system, it is also necessary to consider obstacle avoidance and motion planning.
  Finally, although this study has dealt mainly with the two modalities of vision and body, we would like to extend these modalities in the future.
  We believe that if it becomes possible to handle not only images but also videos, sounds, and tactile sensations in the same way, it will be possible to perform a wider variety of tasks based on linguistic instructions.
  We would like to develop a system that acquires correlations among various sensors and grows autonomously.
}%
{%
  得られた実験結果について考察する.
  まず基礎実験では, 支持した物体方向が一意に定まる統制された環境について, 定量的な実験を行った.
  身体変化と環境変化に基づくネットワークの確率分布変化はParametric Biasの空間に規則的な形で自己組織化可能であった.
  また, これを逐次的に更新することで, 現在の身体や環境の状態を適切に把握することができる.
  そして, 得られたSPNPBを用いることで言語指示に基づき適切な物体の方向に視線を向けることが可能である.
  なお, この性能は確率的な予測モデルとParametric Biasの両方を用いた場合にのみ発揮され, そのうちどちらか一方でも無い場合には, その制御性能は半分程度まで落ちてしまうことが分かった.
  次に応用実験では, より日常生活環境に近い乱雑な環境において実験を行った.
  明るさや人の有無の異なる様々な時間帯においてデータを収集し学習を行うことで, PBの空間にそれら環境変化が自己組織化される.
  また, 基礎実験同様現在の環境を適切に把握することが可能であり, より多様な物体や環境について言語指示から視界制御を行うことが可能であった.
  これらから, 低コスト・低剛性で撓む身体構造であっても, かつ多少の画像変化で出力の変わる大規模視覚-言語モデルを使用した場合でも, 視覚と身体の関係を確率的な形で捉え, 大きな変化をParametric Biasという形で取り込むことで, 適切かつ適応的な視界制御が可能であることが示された.

  本研究の限界と今後の展望について述べる.
  まず, 本研究では主に視界制御について扱ったが, 例えば録画して, という指令に対して実際に録画を実行しているわけではない.
  もちろんそれらは録画という言葉を認識することで容易に実装することができるが, 録画して, や明かりをつけて, だけでなく, 画像をチャットで送って, や書いてある言葉を読み取って, など, その指令には限りがない.
  今後大規模言語モデルを併用し, 言語指令による視界制御と複数のAPIを自動的に利用する, より実用的なシステムへと発展させていきたい.
  次に, もう少し右にして, や奧の方を見て, などの指令は現在受け付けることができていない.
  これは, 大規模視覚-言語モデル自体には身体性が存在しないためであり, 今後の興味深い課題でもある.
  加えて, 定期的にデータ収集と学習を行い, 常に動き続けられるシステム構築も進めていきたい.
  最後に, 本研究では視覚と身体という2つのモーダルを主に扱ったが, これらを今後拡張していきたい.
  静止画像だけでなく動画を用いたり, 音や接触覚等までも同様に扱うことができるようになれば, 言語指示に基づくより多様なタスクが可能になると考える.
  今後より多様なセンサの相関関係を獲得し, 自律的に成長するシステムを構築していきたい.
}%

\section{Conclusion} \label{sec:conclusion}
\switchlanguage%
{%
  This study has described the development of a low-cost low-rigidity robot system that performs daily assistive tasks through view control based on linguistic instructions.
  A neural network is used to learn the correlation between the visual information from CLIP, one of the large-scale vision-language models, and the physical information including target joint angles and necessary joint torques of a low-rigidity robot.
  Here, its probabilistic correlations caused by small changes in the visual field and the low-rigidity body are considered by a predictive model with mean and variance network outputs.
  Changes in the correlations due to changes in time and environment are considered by parametric bias, which is a learnable network input variable.
  The actual robot experiments show that the robot can control its vision from appropriate motion commands according to linguistic instructions, and that open-vocabulary view control is possible even with a low-cost low-rigidity robot.
  In the future, we would like to consider the correlation among language, sound, image, tactile sensation, etc., to enable more advanced robot body control based on linguistic instructions.
}%
{%
  本研究では, 言語指示に基づく視界制御により日常生活支援タスクを行う低コスト・低剛性ロボットのシステム開発について述べた.
  大規模視覚-言語モデルの一つであるCLIPによる視界情報ベクトルと低剛性ロボットの関節角度・関節トルクを含む身体情報の相関関係をニューラルネットワークに基づき学習する.
  この際, 視界の細かな変化や低剛性身体に起因する確率的な相関関係を, 平均と分散のネットワーク出力を持つ予測モデルに基づき捉える.
  また, 時間帯や環境の変化に起因する相関関係の変化を, 学習可能な入力変数であるparametric biasにより捉える.
  これらにより, 言語指示に従って適切な動作指令から視界を制御することが可能となり, 低コスト・低剛性なロボットでもopen-vocabularyなview controlが可能であることを実機実験から示した.
  今後, 言語や音, 画像や接触覚等の間の相関関係を捉え, より高度な言語指示に基づくロボットの身体制御を可能としていきたい.
}%

{
  \bibliographystyle{IEEEtran}
  \bibliography{main}
}

\end{document}